\documentclass[fleqn,10pt]{wlscirep}
\usepackage[utf8]{inputenc}
\usepackage[T1]{fontenc}

\usepackage{blindtext}
\usepackage{amsmath}
\usepackage{amssymb}
\usepackage{lipsum}
\usepackage{graphicx}
\usepackage{etoolbox}
\usepackage{mathtools}
\usepackage{lineno}
\usepackage{hyperref}
\usepackage{float}
\usepackage{authblk}
\usepackage{url}
\usepackage{endnotes}

\title{Contrastive Learning and Mixture of Experts Enables Precise Vector Embeddings}
\author[1]{Logan Hallee}
\author[2]{Rohan Kapur}
\author[3]{Arjun Patel}
\author[4*]{Jason P. Gleghorn}
\author[5*]{Bohdan Khomtchouk}
\affil[1]{{\small Center for Bioinformatics and Computational Biology, University of Delaware}}
\affil[2]{{\small Department of Physics, University of Chicago}}
\affil[3]{{\small The College of the University of Chicago}}
\affil[4]{{\small Department of Biomedical Engineering, University of Delaware}}
\affil[5]{{\small Department of BioHealth Informatics, Luddy School of Informatics, Computing, and Engineering, Indiana University}}
\affil[*]{{\small Correspondence to gleghorn@udel.edu and bokhomt@iu.edu}}


\begin{abstract}
    The advancement of transformer neural networks has significantly elevated the capabilities of sentence similarity models, but they still struggle with highly discriminative tasks and may produce sub-optimal representations of important documents like scientific literature. With the increased reliance on retrieval augmentation and search, representing diverse documents as concise and descriptive vectors is crucial. This paper improves upon the vectors embeddings of scientific text by assembling niche datasets using co-citations as a similarity metric, focusing on biomedical domains. We apply a novel Mixture of Experts (MoE) extension pipeline to pretrained BERT models, where every multi-layer perceptron section is enlarged and copied into multiple distinct experts. Our MoE variants perform well over $N$ scientific domains with $N$ dedicated experts, whereas standard BERT models excel in only one domain at a time. Notably, extending just a single transformer block to MoE captures 85\% of the benefit seen from full MoE extension at every layer. This holds promise for versatile and efficient One-Size-Fits-All transformer networks for numerically representing diverse inputs. Our methodology marks advancements in representation learning and holds promise for enhancing vector database search and compilation.
\end{abstract}
\begin{document}
\flushbottom
\maketitle

\thispagestyle{empty}

\section{Introduction}
The remarkable success of transformer-based large language models (LLMs) since 2017 \cite{NIPS2017_3f5ee243} has significantly increased our confidence in their abilities and outputs. Nowadays, LLMs are treated as a de facto knowledge base for many and have been adopted on a mass scale since the release of services like ChatGPT and open-source counterparts like Llama and Mistral \cite{touvron_llama_2023, jiang_mistral_2023}. However, despite their widespread use, challenges persist, particularly regarding the accuracy and reliability of these models. Common issues like LLM hallucinations \cite{branco-etal-2021-shortcutted,guerreiro-etal-2023-looking} highlight the ongoing need for improvement. The ability to generate reliable vector embeddings and perform precise classification is crucial, especially for technologies that rely on information retrieval and web search.

One approach to further curate transformer latent spaces is to utilize contrastive learning to create sentence similarity models, initially revolutionizing sentiment analysis with broader applications in vector search \cite{reimers-gurevych-2019-sentence, 8642425, https://doi.org/10.1002/cpe.4415}. However, even sentence similarity models miss out-of-distribution domain-specific nuances \cite{giorgi-etal-2021-declutr, Deka_Jurek-Loughrey_Deepak_2021}, resulting in sub-optimal representations of many important documents, including scientific literature.

Fortunately, several advancements have paved the way toward effective sentence similarity models over an arbitrary number of domains. Work from the metascience community has introduced co-citation networks as an easy way to gather many similar papers \cite{ABRISHAMI2019485, 8622810, Brizan_Gallagher_Jahangir_Brown_2016, info13110546, Rodriguez-Prieto_Araujo_Martinez-Romo_2019, ETO2019102046, 10.1145/3209219.3209236, https://doi.org/10.1002/asi.23981, Gipp2009CitationPA}. While this degree of similarity may not be perfect, co-citations have been shown to imply a high degree of similarity between papers \cite{https://doi.org/10.1002/asi.23981}. Another promising advancement comes from the deep learning community with Mixture of Experts (MoE) models. Their learned input-dependent routing of information is a promising multidomain / multitask learning architecture without significant added overhead \cite{Gupta_moemultitask_2022}. Taking advantage of these methods, we propose the following framework to build discriminative vector representations of scientific papers from abstracts alone:

\begin{enumerate}
    \item \textbf{Domain-Specific Fine-Tuning}: Application of contrastive fine-tuning methods to pretrained BERT (Bidirectional Encoder Representation Transformers) models utilizing co-citations as a similarity heuristic, tailoring them to learn and understand specific scientific domains.

    \item \textbf{Universal Applicability through Mixture of Experts (MoE)}: Introduction of a scalable method of seeding MoE models for fine-tuning pretrained BERT models across multiple domains, aiming for a versatile, ``One-Size-Fits-All'' model.
\end{enumerate}

In this paper, we enhance the precision and reliability of LLMs in identifying similar or niche intradisciplinary texts, to build scalable methods that can enhance LLMs to produce effective vector representations from a large variety of scientific literature. Our methods vastly outperform general pretrained models, fine-tuned sentence similarity models, and even science-oriented BERT models. Notably, our MoE variants, equipped with $N$ experts, achieve the efficacy of $N$ individual models, suggesting One-Size-Fits-All transformer networks are possible for certain tasks. Such models have far-reaching implications for information retrieval, web search, and other applications that rely on precise text classification and vector embeddings.

\section{Methods}
\subsection{Data Compilation}
We used co-citation networks to generate sufficiently large and accurate training datasets. Co-citations represent instances where two papers are cited together in a third paper. This strategy enabled the production of large training datasets from small amounts of data due to their nonlinear nature. For a dataset of 10,000 individual papers, for example, over 125,000 co-citation pairs can be produced. While this measurement of similarity is not perfect, co-citations have generally been shown to imply a high degree of similarity between papers \cite{https://doi.org/10.1002/asi.23981}.

Given the technical subfields and the sparse overlap between them, we chose to use the cardiovascular disease (CVD) and chronic obstructive pulmonary disease (COPD) subfields of the biomedical sciences as case studies for our approach. The CVD and COPD subfields represent two domains that contrast significantly in co-citation network size, which allowed us to compare the performance of our approach as it relates to data availability. Since 2010, around 290,000 articles relating to CVD have been published on PubMed Central, while just 14,000 articles relating to COPD were published. We queried papers using Medical Subject Heading (MeSH) terms for both CVD and COPD, specifically for open-access papers with at least one citation and an abstract. We queried specifically for papers published between 2017 - 2022 for CVD and papers published between 2010 - 2022 for COPD. A longer time range for COPD was used as COPD is a smaller sub-field, and a large enough dataset could not be created by querying from 2017 onward. In total, we queried 99,490 papers for our CVD dataset and 10,051 papers for our COPD dataset.

We constructed a test dataset for both sub-fields to prototype our framework on very recent papers, with the goal of applications involving reading list curation for researchers. These were constructed by taking all similar abstract pairs in the training dataset, where at least one paper was published in 2022, the most recent year in our dataset.

Our validation dataset was constructed from the remaining training dataset. The remaining training dataset was split randomly in a 99:1 ratio without duplicates, with the larger of the new datasets being used as our final training dataset and the smaller of the two being used as the initial validation dataset. On top of this initial validation dataset of similar abstract pairs, an equal amount of dissimilar abstract pairs were added. We generated pairs of dissimilar papers by compiling pairs of papers that had never been co-cited together. While it is impossible to guarantee any two non-co-cited papers will not be cited together in the future we minimized this possibility by requiring papers to be cited individually at least 15 times. While we produced pairs of dissimilar papers for model evaluation, the production of dissimilar paper pairs is not necessary for model training, which we discuss in our objective formulation below. 

After constructing the validation and test datasets, we accounted for co-citation frequency in our training dataset by duplicating co-citation pairs that had been co-cited multiple times in the training dataset. If two papers had been co-cited together five times, for example, this duplication would result in this pair of papers occurring five times in our dataset. This duplication allowed us to weigh pairs that had been co-cited more frequently more heavily than pairs that had been co-cited less frequently.

To further diversify the biomedical data available, we applied the same data compilation pipeline to additional sub-fields involving parasitic diseases, skin cancer, and autoimmune diseases. We prototyped our models using separate experiments with only CVD and COPD and then trained them fully on all five domains.

\subsection{Transformer Neural Networks}
The transformer architecture is adept at sequential processing and is SOTA for NLP tasks \cite{open-llm-leaderboard, clark2018think, zellers2019hellaswag, hendrycks2021measuring, lin2022truthfulqa, DBLP:journals/corr/abs-1907-10641, DBLP:journals/corr/abs-2110-14168}. A transformer block comprised a self-attention layer and multi-layer perception (MLP) interleaved with skip connections. Full transformers were made of $T$ transformer blocks stacked together \cite{NIPS2017_3f5ee243}.

Prior to the transformer blocks is the token embedding process, where tokenization maps an input string of language into a list of $L$ integers from a dictionary. These integers served as the indices for a matrix $W_e$, where each row was a learnable representative vector for that token, making $W_e\in\mathbb{R}^{v\times d}$ where $v$ was the total number of unique tokens in the vocabulary and $d$ an arbitrarily chosen hidden dimension. The initial embedding was $\mathbb{R}^{L\times d}$.

Each block in the transformer then \underline{transforms} this embedding, i.e., the $i^{th}$ transformer block maps the embedding $X^{(i-1)} = [x_1^{(i-1)}, ..., x_L^{(i-1)}]^\top \in \mathbb{R}^{L\times d}$ to $X^{(i)} = [x_1^{(i)}, ..., x_L^{(i)}]^\top\in \mathbb{R}^{L \times d}$ \cite{sharma_truth_2023, NIPS2017_3f5ee243, hallee_protein-protein_2023}; $X^{(T)}$ is the last hidden state of the network. The first part of this map is self-attention, which mixes information across the vectors, followed by the MLP which mixes information across $d$ \cite{fu_monarch_2023, sharma_truth_2023}.

Including the MLP, the entire transformer block can be written as
\begin{align*}
    X^{(i)} = \sigma(\text{Attention}(X^{(i-1)})W_1 + b_1)W_2 + b_2,
\end{align*}

where $b_1$ and $b_2$ are biases associated with learned linear transformations $W_1 \in \mathbb{R}^{d\times I}$ and $W_2 \in \mathbb{R}^{I\times d}$, where $I > d$. The activation function $\sigma$, e.g., ReLU or GeLU, introduces non-linearity \cite{NIPS2017_3f5ee243}.

GPT (Generative Pretrained Transformers) models or causal models, like OpenAI's GPT series (GPT-2, GPT-3, etc.), focus on generative tasks and use a variant called transformer decoders \cite{radford2019language, brown_language_2020, openai_gpt-4_2023}. They use unidirectional attention when processing text. This means they can predict the next word in a sentence but cannot modify their understanding based on words that come later in the text. BERT models or transformer encoders utilize bidirectional attention, capture more context and word relationships, and are better suited for tasks like text classification and sentence similarity \cite{devlin_bert_2019}.

\subsubsection{Mixture of Experts}

Mixture of Experts (MoE) models add a linear layer or router network to each transformer block, which outputs logits from $H^{(i)}$. These logits route $H^{(i)}$ to multiple equivalent copies of the MLP section with different weights called experts \cite{shazeer_outrageously_2017}. In many transformer variants, this routing is typically done on a per-token basis, allowing for experts to specify in language classes like punctuation, nouns, numbers, etc \cite{ai_mixtral_2023}. We chose sentence-wise routing of the entire $H^{(i)}$ so that we could purposely structure our experts for specific domains \cite{zuo_moebert_2022}. While there are many ways to route MoE networks, two main approaches involve calling one expert per block or the top $k$ experts per block. We experiment with both approaches.

Controlling the routing of  $H^{(i)}$, allowed for a one-size-fits-all approach to text classification where one expert per transformer layer was an expert in a specific domain. To control this routing, we added special tokens for each domain, like [CVD] and [COPD], and replaced the standard [CLS] token with these upon tokenization. An additional cross-entropy loss was added that compared the router logits to the correct domain identity.

For faster fine-tuning, we utilized pretrained models for this novel \textbf{MoE extension} approach. Our MoE extension took the MLP sections of pretrained transformers and copied them into $E$ experts with randomly initialized routers in each transformer block. In this process, an additional linear layer $W_3 \in \mathbb{R}^{d\times I}$ and bias $b_3$ was added with element-wise multiplication $\odot$ to the MLP (SwiGLU activation) which has been shown to perform better than vanilla activation functions \cite{ai_mixtral_2023}.

\begin{align*}
    Y = (\sigma(XW_1 + b_1) \odot (XW_3 + b_3))W_2 + b_2.
\end{align*}

\noindent We initialized $W_3$ with zeros and $b_3$ with ones to make the initial forward passes equivalent to the pretrained model and would only be modified during further training.

Enforced routing refers to manual indexing of chosen experts, which we found worked just as well as an additional cross-entropy loss on the router logits. We chose enforced routing for the dual CVD / COPD experiments as a proof of concept (we also did not add the additional router linear layer). However, for the experiments over all five biomedical domains, we implemented the mutual information loss suggested in \cite{Chen_mod_2022} to further leverage overlapping similarities across the gathered biomedical domains. This way, we could correlate expert activation with certain domains without direct enforcement and concatenate the top-2 expert results at each layer. Additional local experiments showed that token-wise routing results in slightly higher-end performance, even on sentence-level tasks. Thus, we use the top-2 experts per token for our final five-domain model. Because our MoE extension can be costly in terms of VRAM, we also tried an MoE extension approach with a single transformer block in the middle instead of extending all 12 - hypothesizing that much of the multidomain benefit could be achieved for a small amount of extended MoE layers.

\subsubsection{Models of choice}
We chose two differently sized models to prototype our MoE extension. 1.) all-MiniLM-L6-v2 model (Mini) \cite{wang_minilm_2020, lewis_paq_2021, khashabi_gooaq_2021, dunn_searchqa_2017, koupaee_wikihow_2018, henderson_repository_2019}, a popular sentence transformer model trained on 1+ billion sentence pairs that is highly effective for sentiment analysis and text classification. Mini is a standard BERT-like model with a hidden size of 384, an intermediate size of 1536, 12 attention heads, and six hidden layers for a total of 23 million parameters. 2.) SciBERT is a general-purpose BERT model fine-tuned on entire papers from the Semantic Scholar with over 1.14 million diverse papers \cite{beltagy_scibert_2019}. SciBERT is SOTA for many domain-specific relation extraction and sentence classification tasks with a hidden size of 786, an intermediate size of 3072, 12 attention heads, and 12 hidden layers for a total of 110 million parameters. Compared to many newer BERT architectures with up to (or more than) 1 billion parameters, these modest model sizes allowed for effective fine-tuning given minimal training data and reduced computational cost during training iterations.

MoE versions for the CVD / COPD experiments have a larger parameter count, 30 million and 167 million, respectively, due to two experts (two domains) per layer. However, the effective parameter count is the same as the original because only one expert is called at a time, resulting in fast training and inference. The full domain SciBERT MoE extension versions have five experts and utilize two experts per token for a total of 167 million effective parameters.

Our fine-tuned models were compared to a basic term frequency-inverse document frequency (TF-IDF) model alongside the following popular base models without any fine-tuning: Mini, BERT, Mpnet \cite{song_mpnet_2020}, Declustr \cite{giorgi_declutr_2021}, SciBERT, BiomedBERT \cite{gu_domain-specific_2022}, and ClincalBERT \cite{huang_clinicalbert_2020}. This large model variety in pretraining strategy allows for a more general comparison of how effective our fine-tuning framework is. 

The TF-IDF model, representing the most basic and straightforward sentence similarity model, acts as a baseline for expected performance. Finally, we also prompted GPT3.5 with sets of abstracts to have them assess qualitatively if the papers were similar or not.

\subsection{Training Strategy}

To minimize training time, we chose to use abstracts rather than entire papers as the text input to the model. Abstracts represent a human-generated summarized version of a paper and, as a result, include much of the relevant textual information contained in a paper. We fine-tuned Mini and SciBERT on the individual datasets for CVD and COPD, and SciBERT for all five domains. Models without MoE extension are referred to as Single Expert (SE).

To train our models, we fed co-cited abstracts from the CVD or COPD domain to the model independently, extracting the vector embedding from the pooler output for each abstract. The pooler output was generated by a small neural network from the [CLS] token (our special domain tokens in our case) embedding from the last hidden state $H^{(L)}$, a standard practice when using the Huggingface transformers package. These vector embeddings were compared with a variant of the Multiple Negative Rankings (MNR) loss used to train cdsBERT \cite{hallee_cdsbert_2023, Shariatnia_Simple_CLIP_2021}. MNR Loss is a loss function that has seen significant success with sentence embedding problems \cite{Henderson2017EfficientNL} and was highly successful in our local experiments. Our variant used dot products as a similarity heuristic and scaled the similarity by a learned temperature parameter. Furthermore, MNR loss only requires positive/similar text pairs, generating negative/dissimilar text pairs from the other positive pairs in a given mini-batch. As a result, MNR loss removed the need to generate dissimilar text pairs for our training dataset under the assumption that the random chance of finding a similar paper randomly with a batch size of 20 is sufficiently small. During training, we randomly switched the order of the two input abstract pairs to prevent any bias in how they are fed to the loss function.

We performed hyperparameter optimization on the CVD dataset, using random search to approximate the best batch size alongside warmup steps and total training length that optimized the model's F1 score on the validation dataset. Due to resource constraints, we decided to use a smaller dataset than the actual training dataset, with the hyperparameter optimization dataset being a random 10\% sample of the training dataset. Resource constraints also limited our tested range for each hyperparameter. After trying batch sizes ranging from 5-20, we found that 20 supported the best model performance. This offered a large enough batch size to require nuanced understanding but not too large to hinder model training, as the contrastive loss chosen was significantly more challenging to minimize as the batch size increased. For our final training runs, we utilized a learning rate of $1e^{-5}$, a one-cycle learning rate scheduler \cite{smith_super-convergence_2018} with 500 warmup steps, and periodic validation every 5000 steps. For the final five-domain model, a cosine learning rate scheduler was utilized instead. Training was conducted for ten epochs and halted early when a patience of three was exceeded for the validation $F1_{max}$. The best model was loaded at the end. SciBERT training was conducted on a single A100 GPU, Mini training was done on a single A10 GPU, and all training runs took less than 24 hours.

\subsection{Evaluation Strategy}

\begin{figure}
\includegraphics[width=9cm]{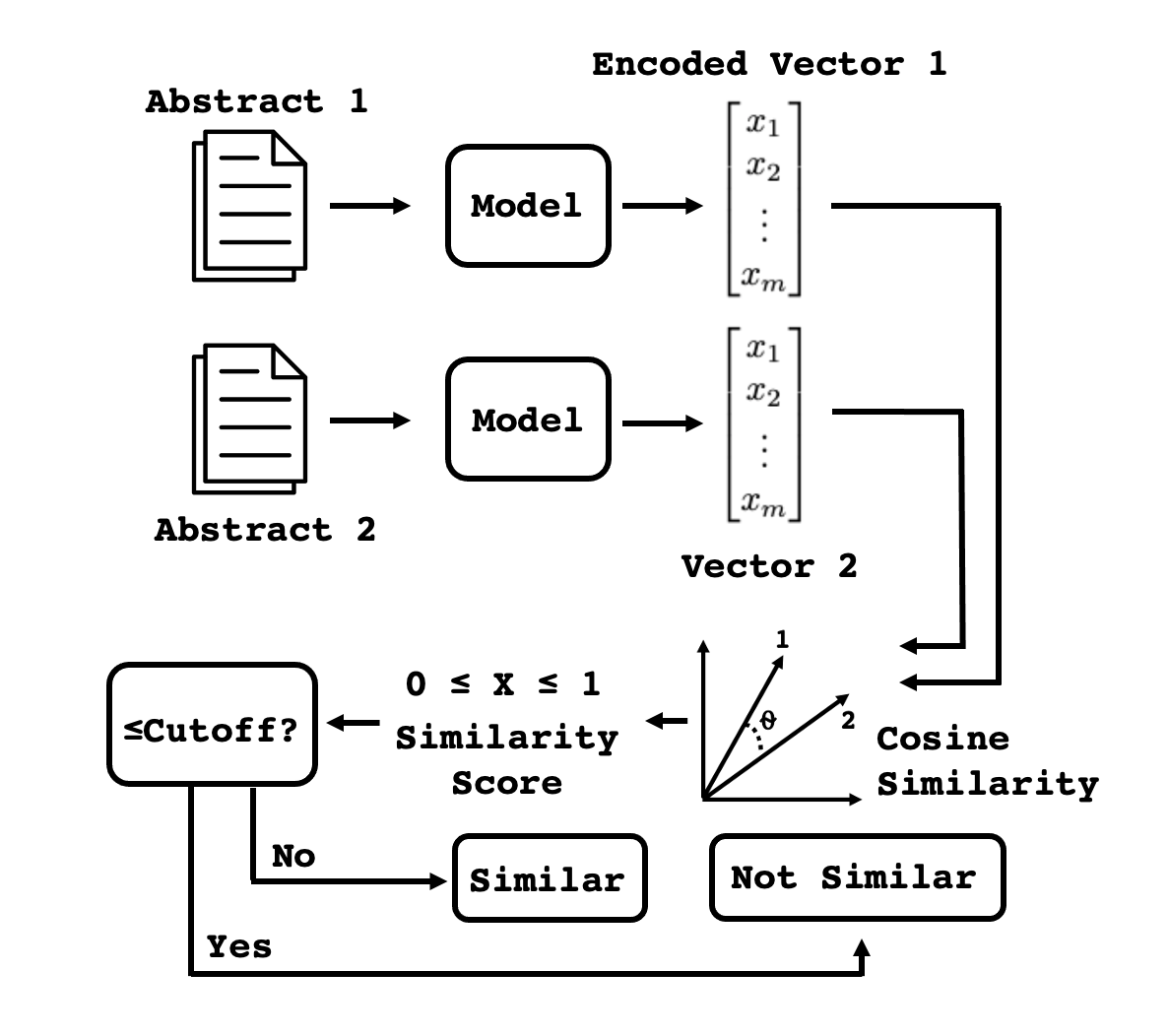} 
\centering
\caption{Method for determination of abstract pair similarity for model evaluation.}
\label{Scoring}
\end{figure}

All models were evaluated on the evaluation sets separately for each domain. We utilized cosine similarity between two vectors extracted from an abstract pair to classify the abstracts as co-cited (similar) or not, given a threshold. Cosine similarity is a common vector similarity measure ranging from -1 to 1, where -1 is exactly opposite and 1 occurs for a pair of the same vector. This binary threshold is determined by an F1 variant called $F1_{max}$ \cite{noauthor_deepgraphlearningtorchdrug_2024, su_saprot_2023}. $F1_{max}$ is the maximum F1 score calculated for all possible thresholds for a reported metric. While typically used for imbalanced multilabel classification, randomly choosing a binary threshold for the reported F1 would not be a fair comparison of different models. For example, one model may perform much better with a cosine similarity threshold at 0.5 for abstract text similarity compared to 0.4, or vice versa. We also reported average distance and accuracy. The average distance was calculated by taking the absolute value of the difference between the similarity score between two abstract pairs and their label, 0 or 1. For example, if a model gave two co-cited (similar) abstracts a cosine similarity of 0.75, the distance would be $(|1-0.75|)$ = 0.25. Full details for conversion of any two abstract pairs into binary similar and dissimilar metrics are summarized in \textbf{Figure \ref{Scoring}}. Because the test dataset contains no negative examples, we limited the $F1_{max}$ threshold search between 0.5 and 1 to prevent the trivial -1 threshold that always leads to $F1_{max}=1$ and reported the accuracy using the found threshold.

\section{Results}
\begin{figure}
\includegraphics[width=7cm]{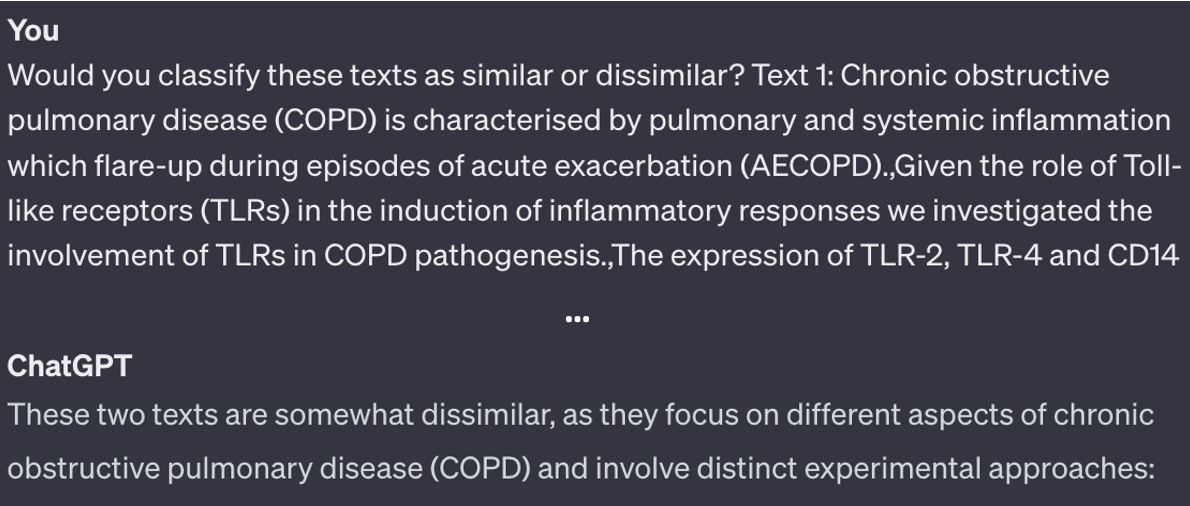} 
\centering
\caption{A typical ChatGPT response to a set of similar papers, qualitatively classifying all similar papers as dissimilar.}
\label{gpt-response}
\end{figure}

The performance of our models on the validation and test datasets for CVD (\textbf{Table \ref{table-cvd}}) and COPD (\textbf{Table \ref{table-copd}}) were summarized and compared to other leading sentence similarity models, as well as TF-IDF. We also evaluated the performance of GPT-3.5 on our generated validation and test datasets. GPT-3.5 represented an example of an LLM with a narrow representation of similar (\textbf{Figure \ref{gpt-response}}). GPT-3.5 regarded almost all examples that were input as dissimilar when asked whether the input examples were similar due to (correctly) identified minor differences between each input text. Given this, GPT-3.5 performance was excluded from model comparisons. We also do not show results from the MoE version of Mini as they did not score better than random on either validation dataset (0.67 $F1_{max}$).

\begin{table}
\centering
\begin{tabular}{||c c c c c c c||}
\hline
  \textbf{CVD Model} & Cutoff & F1Max$_{Valid}\uparrow$ & Dist.$_{Valid}\downarrow$ & Acc.$_{Valid}\uparrow$ & Dist.$_{Test}\downarrow$ & Acc.$_{Test}\uparrow$\\ [0.5ex]
\hline\hline
\multicolumn{7}{||c||}{Our models} \\
\hline\hline
Mini-SE & 0.51 &{0.94} & \textbf{0.23} & {0.93} & 0.19 & 0.97 \\
SciBERT-SE & 0.98 & \textbf{0.97} & 0.48 & \textbf{0.97} & \textbf{0.01} & \textbf{1.00} \\
SciBERT-MoE & 0.98 & \underline{0.97} & 0.48 & \underline{0.97} & \underline{0.01} & \textbf{1.00} \\
\hline\hline
\multicolumn{7}{||c||}{Base models} \\
\hline\hline
TF-IDF & 0.00 & 0.67 & 0.50 & 0.50 & 0.02 & \textbf{1.00} \\
Mini & 0.46 & 0.90 & 0.33 & 0.90 & 0.32 & 0.95 \\
BERT & 0.89 & {0.71}  & 0.48 & 0.65 & 0.07 & 0.89 \\
Mpnet & 0.49  & {0.94}& \underline{0.28} &  {0.94} & 0.26 & 0.96 \\
Declutr & 0.65 & 0.84  & 0.42 & {0.83} & 0.25 & 0.89 \\
SciBERT & 0.44 & 0.89  & 0.34 & 0.90 & 0.39 & 0.87 \\
BiomedBERT & 0.99 & 0.72 & 0.50 & 0.73 & 0.01 & 0.76 \\
ClinicalBERT & 0.92 & 0.71 & 0.49 & 0.68 & 0.06 & 0.81 \\
\hline
\end{tabular}
\caption{Evaluation metrics of fine-tuned models compared to base models for the CVD datasets [\textbf{bold} is best and \underline{underlined} is second best]. Single expert (SE) and Mixture of Expert (MoE) models are compared, showcasing near-identical performance despite MoE models exhibiting mastery of multiple datasets. $F1_{max}$ is not shown for the test set due to the trivial -1 threshold resulting in all models performing perfectly due to no negative data. Thus, a high $F1_{max}$ describes performance for the validation data, and a low distance (Dist) is the important metric for performance on test data.}
\label{table-cvd}
\end{table}

\begin{table}
\centering
\begin{tabular}{||c c c c c c c||}
\hline
  \textbf{COPD Model} & Cutoff & F1Max$_{Valid}\uparrow$ & Dist.$_{Valid}\downarrow$ & Acc.$_{Valid}\uparrow$ & Dist.$_{Test}\downarrow$ & Acc.$_{Test}\uparrow$\\ [0.5ex]
\hline\hline
\multicolumn{7}{||c||}{Our models} \\
\hline\hline
Mini-SE & 0.36 & \textbf{0.83} & \textbf{0.33} & \textbf{0.82} & 0.47 & 0.56 \\
SciBERT-SE & 0.99 & \underline{0.81} & 0.49  & \underline{0.80} & \textbf{0.01} & \textbf{1.00} \\
SciBERT-MoE & 0.98 & 0.80 & 0.49 & 0.79 & \underline{0.01} & \textbf{1.00} \\
\hline\hline
\multicolumn{7}{||c||}{Base models} \\
\hline\hline
TF-IDF & 0.00 & 0.67 & 0.50 & 0.50 & 0.02 & \textbf{1.00} \\
Mini & 0.48 & {0.69} & {0.45} & 0.61 & 0.42 & 0.76 \\
BERT & 0.87 & {0.69}  & 0.49 & 0.58 & 0.09 & 0.85 \\
Mpnet & 0.53 & 0.68  & \underline{0.45}& {0.62} & 0.38 & 0.71 \\
Declutr & 0.62 & {0.68}& 0.47 & {0.56} & 0.27 & 0.91 \\
SciBERT & 0.46 & 0.67  & 0.47 & 0.55 & 0.42 & 0.82 \\
BiomedBERT & 0.98 & 0.67 & 0.50 & 0.51 & 0.01 & 0.97 \\
ClinicalBERT & 0.91 & 0.69 & 0.49 & 0.61 & 0.07 & 0.80 \\
\hline
\end{tabular}
\caption{Evaluation metrics of fine-tuned models compared to base models for the COPD datasets [\textbf{bold} is best and \underline{underlined} is second best]. Single expert (SE) and Mixture of Expert (MoE) models are compared, showcasing near-identical performance despite MoE models exhibiting mastery of multiple datasets. $F1_{max}$ is not shown for the test set due to the trivial -1 threshold resulting in all models performing perfectly due to no negative data.n Thus, a high $F1_{max}$ describes performance for the validation data, and a low distance (Dist) is the important metric for performance on test data.}
\label{table-copd}
\end{table}

\begin{table}
\centering
\begin{tabular}{||c c c c c ||}
\hline
Model & Prec.$\uparrow$ & Recall$\uparrow$ & F1$\uparrow$ & Cutoff\\
\hline\hline
\multicolumn{5}{||c||}{CVD} \\
\hline\hline
SE      & 0.97 & 0.94 & 0.95 & 1.00 \\
MoE     & 0.94 & 0.94 & 0.94 & 0.99 \\
SMoE    & 0.91 & 0.97 & 0.94 & 0.98 \\
SE-All  & 0.55 & 1.00 & 0.71 & 0.99 \\
\hline\hline
\multicolumn{5}{||c||}{COPD} \\
\hline\hline
SE      & 0.74 & 0.88 & 0.80 & 0.99 \\
MoE     & 0.73 & 0.80 & 0.76 & 0.99 \\
SMoE    & 0.58 & 0.98 & 0.73 & 0.98 \\
SE-All  & 0.58 & 0.95 & 0.72 & 0.99 \\
\hline\hline
\multicolumn{5}{||c||}{Skin Cancer} \\
\hline\hline
SE      & 0.72 & 0.88 & 0.79 & 0.98 \\
MoE     & 0.66 & 0.90 & 0.76 & 0.99 \\
SMoE    & 0.58 & 0.96 & 0.73 & 0.98 \\
SE-All  & 0.50 & 1.00 & 0.67 & 0.99 \\
\hline\hline
\multicolumn{5}{||c||}{Autoimmune Disease} \\
\hline\hline
SE      & 0.88 & 0.90 & 0.89 & 0.99 \\
MoE     & 0.86 & 0.92 & 0.89 & 0.99 \\
SMoE    & 0.86 & 0.91 & 0.88 & 0.98 \\
SE-All  & 0.57 & 1.00 & 0.73 & 0.99 \\
\hline\hline
\multicolumn{5}{||c||}{Parasitic Disease} \\
\hline\hline
SE      & 0.88 & 0.93 & 0.90 & 0.99 \\
MoE     & 0.86 & 0.93 & 0.89 & 0.98 \\
SMoE    & 0.89 & 0.90 & 0.89 & 0.99 \\
SE-All  & 0.94 & 0.68 & 0.79 & 1.00 \\
\hline\hline
\multicolumn{5}{||c||}{Average} \\
\hline\hline
SE      & 0.84 & 0.91 & 0.87 & 0.99 \\
MoE     & 0.81 & 0.90 & 0.85 & 0.99 \\
SMoE    & 0.76 & 0.94 & 0.83 & 0.98 \\
SE-All  & 0.63 & 0.93 & 0.72 & 0.99 \\
\hline
\end{tabular}
\caption{Validation metrics across five gathered biomedical domains. SE is trained and evaluated on each domain independently. MoE, SMoE, and SE-All are trained and evaluated on all domains.}
\label{table-full}
\end{table}

The results in \textbf{Tables \ref{table-cvd}} and \textbf{\ref{table-copd}} highlight the effectiveness of our fine-tuning strategy, as our models demonstrated a pronounced proficiency in identifying similar or dissimilar papers within highly specific domains. On the CVD dataset, our models achieved superior $F1_{max}$ and accuracy scores compared to all base models. Particularly, our SciBert variants exhibited a near-perfect 0.97 $F1_{max}$. While the $F1_{max}$ and accuracy scores were lower on the COPD dataset, our models performed better than the base models for the COPD dataset. All our models surpassed every base model evaluated on the COPD datasets by at least 10\% in both $F1_{max}$ and accuracy scores. This performance highlights the capability of our approach to yield high-quality results even with limited training data.

We moved beyond enforced routing and utilized a mutual information loss to train SciBERT MoE extended models across CVD, COPD, skin cancer, autoimmune disease, and parasitic disease domains (\textbf{Table \ref{table-full}}). As a baseline, we trained a model SE-All, which is trained and evaluated on all five domains without MoE extension - effectively fine-tuned SciBERT. Similarly to our initial studies, the MoE extended models performed almost equivalently to five independently trained models together on average, with $0.85$ vs. $0.87$ $F1_{max}$, respectively. The single MoE extended (SMoE) captured 85\% of the added multi-domain proficiency over SE-All with $0.83$ vs. $0.72$ average $F1_{max}$. This ability points to the robustness and versatility of the multidomain / multitask MoE approach through various routing and extension strategies.

\section{Discussion}
Our work advances the use of transformer language models by focusing on improving their domain-specific understanding and document-wide comprehension. We have shown that common pretrained models, including ChatGPT, cannot distinguish the differences in highly discriminative text inputs. Presumably, this phenotype in GPT-like LLMs is primarily caused by standard prompt construction and formatting. By including multiple distinct documents in a single prompt, the semantic similarity of similar tokens will prevent effective distinction between the documents, as portions of each document will attend highly to each other even if they are ``different'' as defined by a desired discrimination. With pretrained BERT-like model usage, presumably this poor performance comes from out-of-distribution tasks and a low ability to generalize with high discrimination. The sentence similarity approach offers a more effective summarization technique by inputting documents separately and allowing the model to construct descriptive numerical representations through contrastive objectives. As vector databases become increasingly prevalent for search and retrieval, the quality of these numerical representations becomes increasingly important. \textit{Our innovative framework, which incorporates contrastive learning, custom MNR and router losses, novel special tokens, MoE seeding, and extension, as well as various routing techniques, significantly enhances sentence similarity metrics compared to pretrained transformers.} We leveraged co-citation networks to construct large datasets of similar abstracts and applied our framework to scientific literature, creating nuanced representations for discriminative subfields of biomedical domains.

Without applying a threshold search for the F1 scores, all base pretrained models tested perform with random chance or worse when tasked with classifying papers as co-cited or not based on cosine similarity. By searching all possible cosine similarity thresholds for $F1_{max}$, pretrained models get a boost for fair comparison while creating a scenario that is unrealistic for actual inference, as a consistent similarity threshold is needed for actual use. Once compared fairly, our fine-tuned models performed better than their original counterparts. More specifically, our SciBERT SE and MoE variants performed equally on the CVD validation set with an $F1_{max}$ of 0.97, an average distance of 0.48, and an accuracy of 0.97. The Mini SE variant performed similarly with a $F1_{max}$ of 0.94, an average distance of 0.23, and an accuracy of 0.93. On the COPD validation set, Mini SE performed best with a $F1_{max}$ of 0.83, an average distance of 0.33, and an accuracy of 0.82. Following that is SciBERT SE with $F1_{max}$ 0.81 and SciBERT MoE with $F1_{max}$ 0.80. Both SciBERT variants resulted in an average distance of 0.49 and accuracies of 0.80 and 0.79, respectively. Importantly, our SciBERT variants consistently performed optimally with a high cosine similarity, implying that a standardized threshold near 0.98 or 0.99 could be utilized during inference. Further penalization terms on the average similarity between batches during training could enforce this threshold lower for different applications.

The test sets require closer examination due to the lack of negative examples, constrained by the inclusion of newer papers and a lack of literature cited 15+ times since 2022. To accommodate this, we withheld $F1_{max}$ from the tables since there is a trivial cutoff of -1 to get 1.00 $F1_{max}$. Instead, we limited the threshold search between 0.5 and 1.0 and reported the accuracy using the found threshold, providing a nontrivial representation of performance. Overall, we found that SciBERT variants demonstrated more precise vector representations with 1.00 accuracy across the board. In contrast, Mini SE variants had lower accuracies of 0.97 for CVD and 0.56 for COPD. Surprisingly, the Mini SE \textit{CVD variant} performed better than the Mini SE COPD variant on the \textit{COPD test data}. This suggests that the cosine similarity threshold limit of 0.5 may have artificially hindered the metrics evaluated. The average distance metric offers additional context for test dataset performance. SciBERT variants excelled at placing similar abstract vector representations close in space with an average distance of 0.01 for CVD and COPD test sets. Conversely, the smaller Mini SE had a high distance even compared to base models. Notably, the BiomedBERT base model also had an average test set distance of 0.01 on both test sets, which is unsurprising given the possibility of training data overlap. Despite MoE engineering, our attempts with a Mini MoE variant were less successful, suggesting a minimum size requirement in the base BERT model for effective performance. This may be due to the need for sufficiently capable shared attention layers that can generalize to support diverse experts and domains.

Importantly, our MoE approach across all domains performs similarly to our individual-domain SE models. Additionally, the MoE seeding is fully scalable, appearing to enable $N$ experts and $N$ new special tokens given $N$ different datasets. This is further supported by our experiments shown in \textbf{Table \ref{table-full}}, where MoE models with five experts perform well on five domains, even with a single MoE layer. The substantial improvement in performance achieved by adding a single MoE layer highlights a remarkable benefit with little added computational cost. This is particularly promising for using our single-layer MoE extension with large pretrained models to train for multitask / multidomain tasks. Future experiments may find the optimal ratio of experts per domain alongside the correct discrimination of ``domain'' to create a one-size-fits-all vector embedding model at the scale of Semantic Scholar.

Our use of co-citation networks enables rapid and efficient dataset compilation for training transformers in niche scientific domains. Fine-tuning of base BERT models through contrastive learning with an MNR-inspired loss significantly improves sentence similarity. The MoE approach further expands these capabilities, suggesting the feasibility of a universal model for text classification and vector embeddings across various domains through MoE seeding and enforced routing. Using these new models, effective MoE BERT models with specialized knowledge across multiple fields, vocabularies, or tasks can be developed.

\subsection*{Data and Code Availability}
\small Links to all training data, model weights, and code can be found at \href{https://github.com/Gleghorn-Lab/Mixture-of-Experts-Sentence-Similarity}{Github}.

\subsection*{Funding and Acknowledgments} 
\small The authors thank Katherine M. Nelson, Ph.D., for reviewing and commenting on drafts of the manuscript. This work was partly supported by grants from the University of Delaware Graduate College through the Unidel Distinguished Graduate Scholar Award (LH), the National Institutes of Health R01HL133163 (JPG), National Institutes of Health R01HL145147 (JPG), Indiana University (BBK), and NIH R01DK132090 (BBK).

\subsection*{Contributions}
\small Conceptualization (RK, AP, BBK, LH, JPG), Co-citation Methodology (RK, AP, BBK), MoE Methodology (LH, JPG), Data Curation (RK, AP, BBK), Investigation (LH, RK, AP, BBK), Formal Analysis (LH, RK, AP, JPG, BBK), Writing – Original Draft (LH, RK, AP, BBK), Writing – Review \& Editing (LH, JPG, BBK), Supervision (JPG, BBK), Project Administration (JPG, BBK), Funding acquisition (LH, JPG, BBK).

\subsection*{Conflict of interest}
\small The authors declare no conflict of interest.

{\footnotesize\bibliography{references}}

\end{document}